\documentclass[10pt, conference, a4paper]{IEEEtran}
\usepackage[utf8]{inputenc}

\usepackage{microtype}
\usepackage{amsmath}
\usepackage{amsfonts} 
\usepackage{relsize}
\usepackage{graphicx}
\usepackage{bbm}
\usepackage{enumitem}
\usepackage{dblfloatfix}
\usepackage{svg}

\title{Towards Interpretable Deep Reinforcement Learning Models via Inverse Reinforcement Learning}
\author{
\IEEEauthorblockN{Sean Xie}
\IEEEauthorblockA{Department of Computer Science\\
Dartmouth College\\
Hanover, NH 03755\\
yuansheng.xie.gr@dartmouth.edu}
\and
\IEEEauthorblockN{Soroush Vosoughi}
\IEEEauthorblockA{Department of Computer Science\\
Dartmouth College\\
Hanover, NH 03755\\
soroush.vosoughi@dartmouth.edu}
\and
\IEEEauthorblockN{Saeed Hassanpour}
\IEEEauthorblockA{Department of Biomedical Data Science\\
Dartmouth College\\
Hanover, NH 03755\\
saeed.hassanpour@dartmouth.edu}
}

\begin{document}
\maketitle

\begin{abstract}
Artificial Intelligence, particularly through recent advancements in deep learning (DL), has achieved exceptional performances in many tasks in fields such as natural language processing and computer vision. For certain high-stake domains, in addition to desirable performance metrics, a high level of interpretability is often required in order for AI to be reliably utilized.  Unfortunately, the black box nature of DL models prevents researchers from providing explicative descriptions for a DL model’s reasoning process and decisions. In this work, we propose a novel framework utilizing Adversarial Inverse Reinforcement Learning that can provide global explanations for decisions made by a Reinforcement Learning model and capture intuitive tendencies that the model follows by summarizing the model’s decision-making process.
\end{abstract}

\section{INTRODUCTION}
    \IEEEPARstart{T}{he} utilization of deep learning (DL) models has become increasingly prevalent in a variety of domains such as smart city \cite{chen2019survey, https://doi.org/10.48550/arxiv.2206.03132}, criminal justice \cite{dass2022detecting, rigano2019using, rudin2019stop}, drug discovery \cite{vamathevan2019applications, jimenez2020drug} and healthcare \cite{mcbee2018deep, coudray2020deep, dipalma2021resolution}. In these domains, the successful adoption of predictive models in assisting with high-stake decisions depends not only on the performance of these models but also on how well the process by which these models are making their decisions can be understood by their users \cite{ahmad2018interpretable, xie2022interpretation}. Only when users clearly understand the behavior of a model,  can they determine how much they should rely on the a model's decisions. The complexity of DL models, however, make it an extremely difficult task to explain or reason about their behaviors \cite{arrieta2020explainable, ribeiro2016should}. The crux of the problem rests with the basis of DL, the Multi-Layer Perceptron (MLP). Since their inception, MLPs have been widely regarded as being capable of inferring and modeling complex relationships in data. However, due to the structure and the enormous number of parameters in MLPs, MLPs are not by their nature interpretable. The necessity of model interpretability means that the MLP's black-box nature is often an impediment to DL’s ability to generate practical value. Researchers and industry professionals frequently find themselves hesitant to utilize DL models in their applications due to the models' lack of transparency \cite{dovsilovic2018explainable}. As result, external methods that can faithfully explicate the intricacies in a DL model have become a topic of much interest.  The Explainable Artificial Intelligence (XAI) community has recently begun to place more focus in this area, specifically in investigating new interpretability techniques for MLP and deep learning models, including model simplification approaches, feature relevance estimators, text explanations, local explanations and model visualizations \cite{arrieta2020explainable}. The most popular interpretability methods, however, are all local, i.e. they are based on single instances of model decisions and provide explanations for those instances individually. Current global methods, i.e. methods that summarize and describe a model, are often task-specific or not suited for deep learning models with a wide range of features. 
    Inverse Reinforcement Learning (IRL) is a class of methods that seeks to learn the policies of trained models (experts) by inferring the reward function from demonstrations \cite{argall2009survey}. A particular subclass of IRL methods, named Adversarial Inverse Reinforcement Learning (AIRL), learns and encapsulates the reward function in the form of a discriminator that can discern expert trajectories from non-expert trajectories \cite{fu2017learning}. This discriminator is then used as an approximation for the true reward function in order to train a novice agent from scratch. The discriminator captures the training objective and induces expert-like output from the novice agent. Therefore, the discriminator is of interest from the viewpoint of interpretability  because it can be utilized to explain the decision-making process of a model globally. 
    In this work, we introduce a novel interpretability framework to provide explanations for deep reinforcement learning models. This framework leverages AIRL to generate a discriminator and then utilizes the discriminator to discover patterns and tendencies that are latent in the decision-making process of a Deep Reinforcement Learning (DRL) model. Our framework has three main contributions:
    
    \begin{enumerate}
        \item In contrast to existing interpretability methods, our framework provides a comprehensive global summarization of a model without focusing on a singular input.
        \item Our framework, to the best of our knowledge, is the first attempt at using Inverse Reinforcement Learning to explain and interpret abstractive summarization
        \item Our framework is inherently interpretable in that the explanations produced by the framework are easy to understand.
    \end{enumerate}

\section{RELATED WORK}
    The field of Explainable Artificial Intelligence (XAI) is a rapidly evolving field \cite{arrieta2020explainable}. Current methods in XAI that focus on the interpretability of machine learning models can generally be categorized as either global (i.e., summarizing the model and its behavior under different settings) or local (deciphering the process that produced the prediction for a single instance). There also exist three main popular approaches for machine learning models to attribute importance to different parts of the input. 
    
    \textbf{Gradient-based} approaches measure the amount of change around a neighborhood in the model that is necessary to induce a change in the output. The importance of features is determined through derivative calculations with respect to the input of the models \cite{sundararajan2017axiomatic, smilkov2017smoothgrad}. The popularity of this method is due to the fact that the attribution of importance to different features, which naturally lends itself to interpretability, and can be computed conveniently via back propagation \cite{rumelhart1986learning}. Although methods that use this approach are straightforward, they are, by definition, only local. In addition, some methods using this approach can only reproduce a subset of the original features and their corresponding relevance \cite{adebayo2018sanity}.
    
    \textbf{Surrogate-based} approaches \cite{bach2015pixel, ribeiro2016should, shrikumar2017learning, lundberg2017unified, xie2023proto}aim to develop a secondary model that can approximate the primary model but the secondary model is constructed in a way that is more easily interpretable. Additive feature attribution methods, where features are assigned a coefficient and the approximation of the original model is calculated through a linear combination of the features and their corresponding coefficients, is a well-known and prominent surrogate-based method. It should be noted that the most popular interpretability methods used today (e.g. Local Interpretable Model-agnostic Explanations (LIME) ) all utilize some combinations of gradient-based and surrogate-based approaches. Also, they are all local and do not offer a general comprehension of the model itself. Current global methods \cite{DBLP:journals/corr/LakkarajuKCL17, deng2019interpreting} for these two approaches rely on decision trees and decision sets, which become difficult to utilize and interpret as the number of features in the inputs increases and the nature of inputs becomes more abstract as the case in NLP text-generation tasks. 
    
    \textbf{Perturbation-based} approaches \cite{balls1996investigating, vstrumbelj2009explaining, fong2017interpretable, olden2002illuminating} modify certain parts of the input and use detected change in the output to assess the importance of features in the input. Although perturbation-based methods directly estimate feature importance with relative clarity, they become computationally prohibitive as the number of features in the input increases \cite{zintgraf2017visualizing}. In addition, the product of perturbation-based methods is highly contingent upon the number of features that are modified in the process \cite{ancona2017towards}. Finally, perturbation-based methods are usually employed in such a way that they are local to each individual instance. 
    
    \textbf{Our framework} is categorized as global as it does not fixate on a single instance but rather produces results that can generalize and summarize the behavior of the model as a whole. Policy extraction and summarization methods via IRL have been explored by \cite{pan2020xgail, lage2019exploring}, and \cite{amir2018highlights} before. However, none of those works were employed in the NLP domain and utilized the Adversarial method of IRL in their approaches. Our framework is also similar to surrogate-based approaches in that it works by approximating the reward function under which the expert was trained with a discriminator and mines information from this reward function. In contrast to existing global surrogate methods such as \cite{DBLP:journals/corr/LakkarajuKCL17, deng2019interpreting} and \cite{topin2019generation}, the rules and input-output relationships mined by our framework are easily interpretable for users even as the number of features substantially increases.

\section{METHODOLOGY}
    Our three-step framework relies on the techniques of Reinforcement Learning, Inverse Reinforcement Learning, and Adversarial Inverse Reinforcement Learning. The detailed usages of each of them are provided below. 
    
    \subsection{Reinforcement Learning}
    A Markov Decision Process (MDP) \cite{puterman2014markov} is defined as a tuple $M = (S, A, T, r, \gamma, \rho)$, where S and A are, respectively, the state and action spaces, $\gamma \in  (0, 1)$ is the discount factor, $T(s' | a, s)$ is the transition model that defines the conditional distribution of the next state, $s'$, given the current state ($s$) and action ($a$), $r(s,a)$ is the reward function that defines a reward for performing action $a$ while in state s, and $\rho$ is the initial state distribution. Reinforcement Learning seeks to find the optimal policy $\pi^*$, which is defined as the policy that maximizes the discounted sum of rewards \cite{metelli2017compatible}. 
    
    \begin{equation}
        \label{expectation_reinforcement_learning}
        \mathlarger{\mathlarger{\mathlarger{\mathbf{E}}}}  \left[ \sum_{t=0}^{\infty} \gamma^t R(s_t, a_t) \vert \pi \right]
    \end{equation}
    
    \subsection{Inverse Reinforcement Learning (IRL)}
    Under the same MDP $M$ but without the reward function  $r(s, a)$, the goal of IRL is to infer the reward function $r(s, a)$ given a set of demonstration trajectories $D = \{\tau_1, …, \tau_N\}$. We assume these demonstration trajectories are produced by an agent operating under optimal policy $\pi^*$(an expert). Solving the IRL problem can be framed as solving the maximum likelihood problem:
    
    \begin{equation}
        \label{irl_maximum_likelihood}
        \mathop{\mathbf{max}}_{\theta} \mathlarger{\mathlarger{\mathbf{E}}}_{\tau \sim D} \left[ \log p_{\theta}(\tau) \right]
    \end{equation}
    
    where 
    \begin{equation}
        \label{probability_of_tau}
        p_{\theta}(\tau) \propto p(s_0) \prod_{t=0} p(s_{t+1} \vert s_t, a_t)e^{\gamma^t r_{\theta}(s_t, a_t)}
    \end{equation}
    parameterizes the reward function $r_\theta(s, a)$ but fixes the dynamics and initial state distribution to that of $M$ \cite{fu2017learning}.
    
    \subsection{Adversarial Inverse Reinforcement Learning}
    Optimizing (\ref{irl_maximum_likelihood}) can be formulated as a generative adversarial network (GAN) \cite{qureshi2018adversarial} optimization problem with the discriminator taking the form of

    \begin{equation}
        \label{discriminator}
        Disc(s, a, s') = \frac{\exp(f_{\theta, \phi} (s, a, s'))} 
                              {\exp{(f_{\theta, \phi} (s, a, s')} + \pi(a \vert s))}
    \end{equation}
    where $f_{\theta, \phi}$ is a learned function
    comprising of a $g_\theta$ term that approximates the reward and a $h_\phi$
    term that shapes the reward. This particular approach to inverse
    reinforcement learning is named Adversarial Inverse
    Reinforcement Learning (AIRL). The equation for $f_{\theta, \phi}$ is as
    follows:
    
    \begin{equation}
        \label{f_theta_phi}
        f_{\theta, \phi}(s, a, s') = g_\theta(s,a) + \gamma h_\phi(s') - h_\phi(s)
    \end{equation}
    
    The discriminator is trained via binary logistic regression to
    discern expert data trajectories from non-expert (novice)
    trajectories. The output of the discriminator can be viewed as
    the reward for the transition from state s to state s’ by taking
    action a. The discriminator’s output rewards are then used to
    train the novice to be more expert-like. For details on AIRL,
    we refer the reader to \cite{fu2017learning} and \cite{qureshi2018adversarial}.
    
    \subsection{Our Framework}
    We break down our framework (Fig. 1) into three steps:\\ 
    \begin{itemize}
        \item \textbf{Step 1}: The RL Training part of our framework involves
        training a model via reinforcement learning and obtains an
        expert model that operates under $\pi^*$.
        
        \item \textbf{Step 2}: The AIRL Training part of our framework involves
        first initializing an untrained model (which will become the
        novice agent) that has the same model architecture as the
        expert. We then utilize the AIRL algorithm to pit the expert
        and the novice against each other by having the discriminator
        discern trajectories produced by the expert from trajectories
        produced by the novice. The discriminator, defined in (\ref{discriminator})
        by the end of this step will be able to reward expert-like
        behavior by generating a higher numerical output for expertlike trajectories. By the end of this step the novice will also be
        able to produce expert-like trajectories up to a certain level.
        
        \item \textbf{Step 3}: This step of our framework is about analyzing the
        discriminator’s rewards. We define a trajectory $\tau = (s_0, a_0, s_1, a_1 ,…, s_t, a_t)$ as a sequence
        of states and actions produced by a model during its prediction
        process for a single instance. We define $D = \{\tau_1, …, \tau_N\}$ to be trajectories of a model over all
        instances in the dataset. We take trajectories produced by a
        model (in our case, the expert model) and feed them to the
        discriminator to produce outputs for each $\tau \in D$. The outputs of 
        the discriminator over all $D$ can then be and analyzed to
        capture trends and patterns in the discriminator’s reward data.
        Because the discriminator’s output is the reward for
        transitioning from state $s$ to state $s’$ by taking action a, we can
        use this output to rate a model’s decisions and empirically
        compute aggregate information to gain an understanding over
        all the decisions the model has made – thus giving us a global
        summarization of the model’s decision-making process.
        Intuitively, because the novice is trained using only reward
        data from the discriminator, the discriminator’s output should
        provide valuable information regarding how the novice (or any RL model
        with the same architecture as the expert) prefers to transition
        and the underlying decision-making pattern when taking these transitions. In the following sections we
        describe our specific experiment and the methods we used to
        collect and aggregate the necessary information to find patterns
        for our specific dataset and application. Please note that, for a
        different task in a different domain, the reward analysis part
        will need to be adapted to gather appropriate information.
    \end{itemize}
    
    \begin{figure}[!t]
        \centering
        \includegraphics[width=3in]{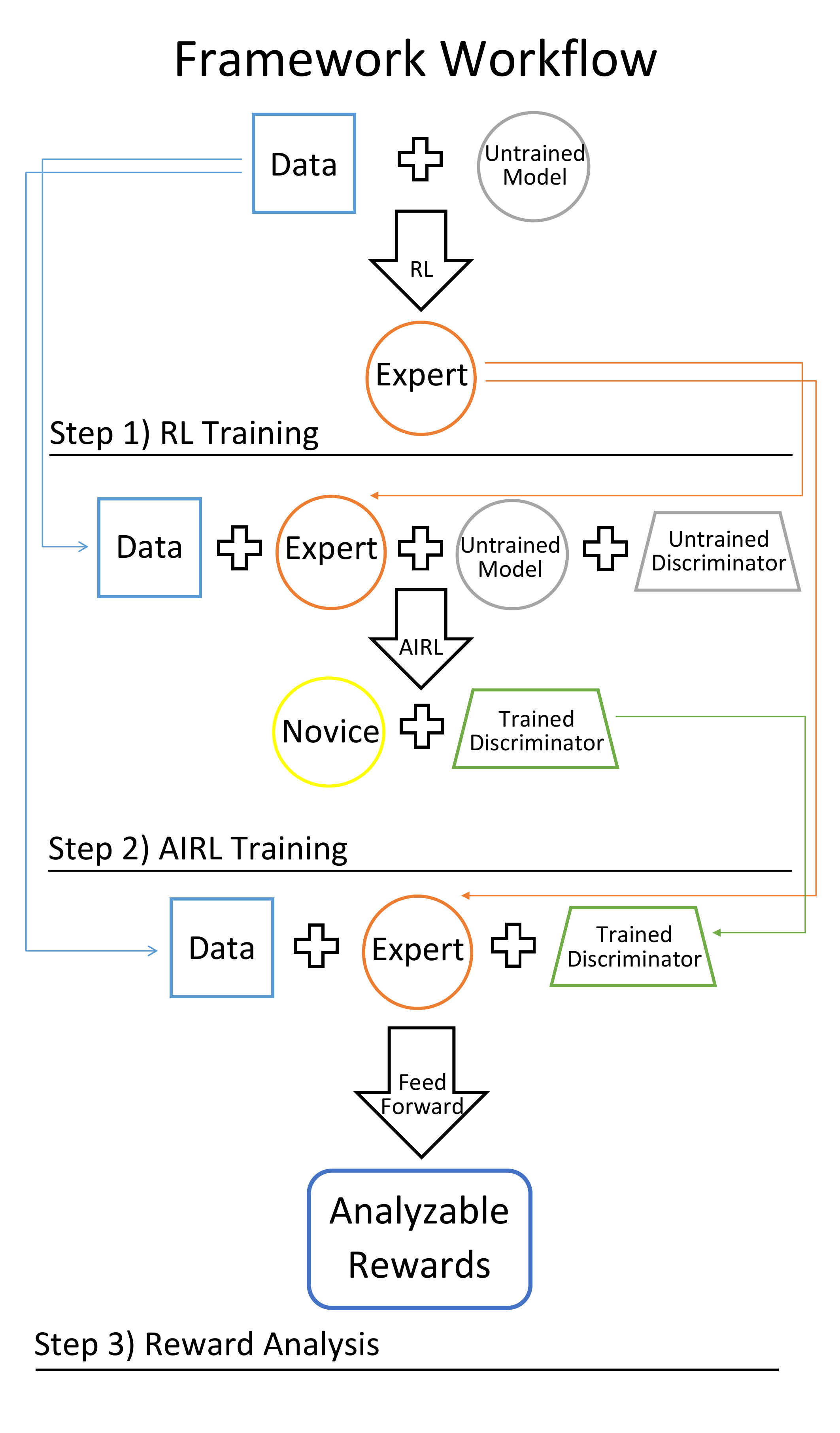}
        \caption{Our 3-step Framework to increase model interpretability}
        \label{framework_figure}
    \end{figure}
    
\section{EXPERIMENT}
    
    \subsection{Dataset}
        The dataset we utilized in our experiment is the CNN/Daily Mail
        dataset introduced by \cite{hermann2015teaching}. A single sample in the dataset
        contains a news article sourced from CNN/Daily Mail and its
        corresponding ground truth summary. 287,113 such samples in
        the dataset were used for training, 13,368 samples for
        validation and 11,490 samples for testing.
        
    \subsection{Our Task}
        We tested our framework on an abstractive summarization
        task. Our expert (denoted here as $\pi^*$ is a self-attentive model
        trained via deep reinforcement learning. Our novice model
        (denoted here as $\pi$) is of the same architecture as the expert but
        trained via AIRL. The models take as input a tokenized news
        article and abstractly produces a summary for that article,
        which we then evaluate against the human-generated ground
        truth summary using ROUGE \cite{lin2004rouge}. We chose abstractive
        summarization in the NLP domain as our representative RL-based task because textual data is interesting in that each unit
        of datum (i.e., each word) has many characteristics to be
        analyzed. In our experiment, for each word, we chose to
        analyze its frequency of appearance, its complexity (length in
        characters), and its part of speech.

    \subsection{Model Architecture and Implementation}
        The architectures of our expert and novice models are both
        encoder-decoder that employs intra-temporal attention in the
        encoder (bi-directional LSTM) and intra-decoder attention in
        the decoder (uni-direction LSTM) \cite{schmidhuber1997long}. These mechanisms
        have been shown by \cite{paulus2017deep} to produce good summaries that
        contain fewer repetitions in a DRL environment. We limit the
        vocabulary to 50,000 words. For each input article, at each time
        step $t$, our model generates probability distribution $P^t$ from which the next word in the summary is sampled. 
        We train our expert model that is fed into the framework first with the loss
        function specified in the next section. For additional details on
        the architecture of the model and hyper parameters used during
        training, we refer the reader to the technical appendix in the
        supplementary materials.
    
    \subsection{Loss Function}
        Our loss function during RL training incorporates the
        ROUGE function as part of the reward. For a given input
        sequence $y^{in}$ in, our models output two sequences, $y^s$ and $y^g$.
        Sequence $y^s$ is the sequence the model produces by sampling
        from the distribution $P^t$ each time step t. Sequence $y^g$
        is the sequence the model produces by greedily selecting the token
        that would maximize the output probability at each time step t.
        We use the ROUGE function as the reward function for output
        sequences and incorporate its output into our loss function.
        Specifically, we define the ground truth summarization for a
        particular sample as $y^G$ and function $R(y')$ as a function that
        returns the ROUGE score between the outputted sequence of
        $y'$ and the ground truth sequence $y^G$. Our RL model’s loss
        function is shown in (\ref{loss_function}). During training, minimizing (\ref{loss_function}) is the same as increasing the reward expectation in (\ref{expectation_reinforcement_learning})
        of our expert model.
    
    \begin{equation}
        \label{loss_function}
        \mathcal{L} = (R(y^g) - R(y^s)) \sum_{t=1}^{N} \log P^t (y^s_t \vert y^s_1, ..., y^s_{t-1}, y^{in})
    \end{equation}
    
    \subsection{Training the Novice and the Discriminator}
        The expert, untrained novice, and untrained discriminator
        are used together in the AIRL algorithm. In our application, the
        two functions $g_\theta$ and $h_\phi$ that make up the discriminator $f_{\theta, \phi}$ 
        are both MLPs. The discriminator takes as input trajectories
        of the expert or the novice and outputs a reward for the input
        trajectories. For the purpose of inputting into the discriminator,
        we reformat $\tau$ as defined in the Methodology section into
        sequences of tuples that represent actions that caused a
        transition of states, i.e., $\tau = ((s_0, a_0, s_1), (s_1, a_1, s_2),…, (s_{T-1}, a_{T-1},
        s_{T}))$. By the end of step 2 in Fig. \ref{framework_figure}, our discriminator can
        discern between expert and non-expert trajectories and
        correspondingly reward expert-like trajectories. We show
        through our experimentation that the novice is able to
        reproduce the expert’s trajectories up to a certain degree. 
    
    \subsection{Aggregate Analysis of Rewards}
        After training our discriminator we run the expert’s trajectories
        through the discriminator and obtain rewards for all trajectories
        in the dataset. We then take the softmax of rewards within each
        singular trajectory to obtain normalized rewards for each
        singular trajectory. For this particular task, to get a general
        description of our model, we calculated the most rewarding
        parts of speech (with two methods) and the normalized mutual
        information (MI) score between word characteristics and
        rewards, which we discuss below
    
\section{RESULTS}
    
    \subsection{Model Evaluation}
    shows a summary of the results of our expert and
    novice models on the testing set. We also include the
    performance of three State-of-the-Art (SOTA) models for
    reference.
    
    \begin{table}[!b]
        \renewcommand{\arraystretch}{1.3}
        \caption{Rouge Scores of Models on CNN/Daily Mil Dataset}
        \label{evaluation_table}
        \centering
        \begin{tabular}{c|c|c|c}
            \hline
            \bfseries Model & \bfseries Rouge-1 F1 & \bfseries Rouge-2 F1 & \bfseries Rouge-l F1 \\
            \hline
            Expert & 42.80 & 19.12 & 27.96 \\
            \hline
            Novice & 14.21 & 8.23 & 8.11 \\
            \hline
            BART \cite{lewis2019bart} & 44.16 & 21.28 & 40.96 \\
            \hline
            T5 \cite{raffel2019exploring} & 43.52 & 21.55 & 40.69 \\
            \hline
            UniLMv2 \cite{bao2020unilmv2} & 43.16 & 20.43 & 40.34 \\
            \hline
            
        \end{tabular}
    \end{table}
    
    \subsection{Most Rewarding Parts of Speech}
        We summed up the rewards by parts of speech over all
        trajectories and calculated the average for each tag to find the
        most rewarding the POS tags. The 36 tags we use here are from
        The Penn Treebank \cite{marcus1994penn}. We treat words that have different
        parts of speech as different words, i.e. (run, noun) and (run,
        verb) are treated as different words. We took the averages
        through two methods. For Method 1, we first grouped the
        rewards by words and divided by the number of times those
        words appeared before dividing by the number of times that
        particular parts of speech appeared. This is done to prevent
        words that appear more frequent from having an unfair impact
        on the rewards. For method 2, we forwent normalizing by the
        number of times each word appeared and instead summed up
        the rewards for each parts of speech over all words with that
        part of speech and divided by the total number of times the
        part of speech appeared. We define the symbols below in
        Table II. and provide equations for the two ways of averaging
        parts of speeches in (\ref{averaging_methods}).
        
    \begin{table}[!t]
        \renewcommand{\arraystretch}{1.3}
        \caption{Definitions of Symbols Used in (\ref{averaging_methods})}
        \label{evaluation_table}
        \centering
        \begin{tabular}{c|c}
            \hline
            \bfseries Symbol & \bfseries Definition \\
            \hline
            $S$ & the set of all parts of speech in the Penn Tree Bank \\
            \hline
            $D$ & the set of all expert trajectories \\
            \hline
            $V$ & the set of all words that appeared in $D$ \\
            \hline
            $n_s$ & number of times a POS $s \in S$ appeared in $D$ \\
            \hline
            $w_s$ & a word that has part of speech $s$\\
            \hline
            $n_{w_{s}}$ & the number of times $w_s$ appeared in D \\
            \hline
            
            $r_{w_{s},\tau}$ & the total reward of $w_s$ in a trajectory $\tau \in D$\\
            \hline
            
        \end{tabular}
    \end{table}
        
    \begin{equation}
        \label{averaging_methods}
        \forall s \in S 
        \begin{cases} 
          \text{Avg. Method 1} = \mathlarger{\frac{1}{n_s}} \mathlarger{\sum_{w_s \in V}} \mathlarger{\frac{1}{n_{w_s}}} \mathlarger{\sum_{\tau \in D} r_{w_s, \tau}} \\
          \text{Avg. Method 2} = \mathlarger{\frac{1}{n_s}} \mathlarger{\sum_{w_s \in V}} \mathlarger{\sum_{\tau \in D} r_{w_s, \tau}} \\
        \end{cases}
    \end{equation}
    
    \subsection{Mutual Information}
            Mutual Information (MI) is a measure of dependency
        between two variables. We calculate the normalized MI score
        between 3 characteristics of words and the discriminator’s
        rewards for transitions involving these words and present them
        in Table III. The 3 characteristics that we analyzed are:
        \begin{itemize}
            \item The number of appearances is the number of times the word has appeared throughout the expertgenerated summaries.
            \item The complexity of a word is the number of letters in the word.
            \item The part of speech of a word. We also use the POSs from the Penn Treebank here. 
        \end{itemize}
        
        \begin{table}[!b]
        \renewcommand{\arraystretch}{1.3}
        \caption{Normalized MI Scores between Word Characteristics and Word Rewards}
        \label{evaluation_table}
        \centering
            \begin{tabular}{c|c}
                \hline
                \bfseries Characteristic & \bfseries MI Score with Rewards \\
                \hline
                Number of Appearances & 0.42 \\
                \hline
                Complexity & 0.24 \\
                \hline
                Part of Speech & 0.2 \\
            \end{tabular}
        \end{table}  
    
    \begin{figure}[!t]
        \centering
        \includegraphics[width=2.75in]{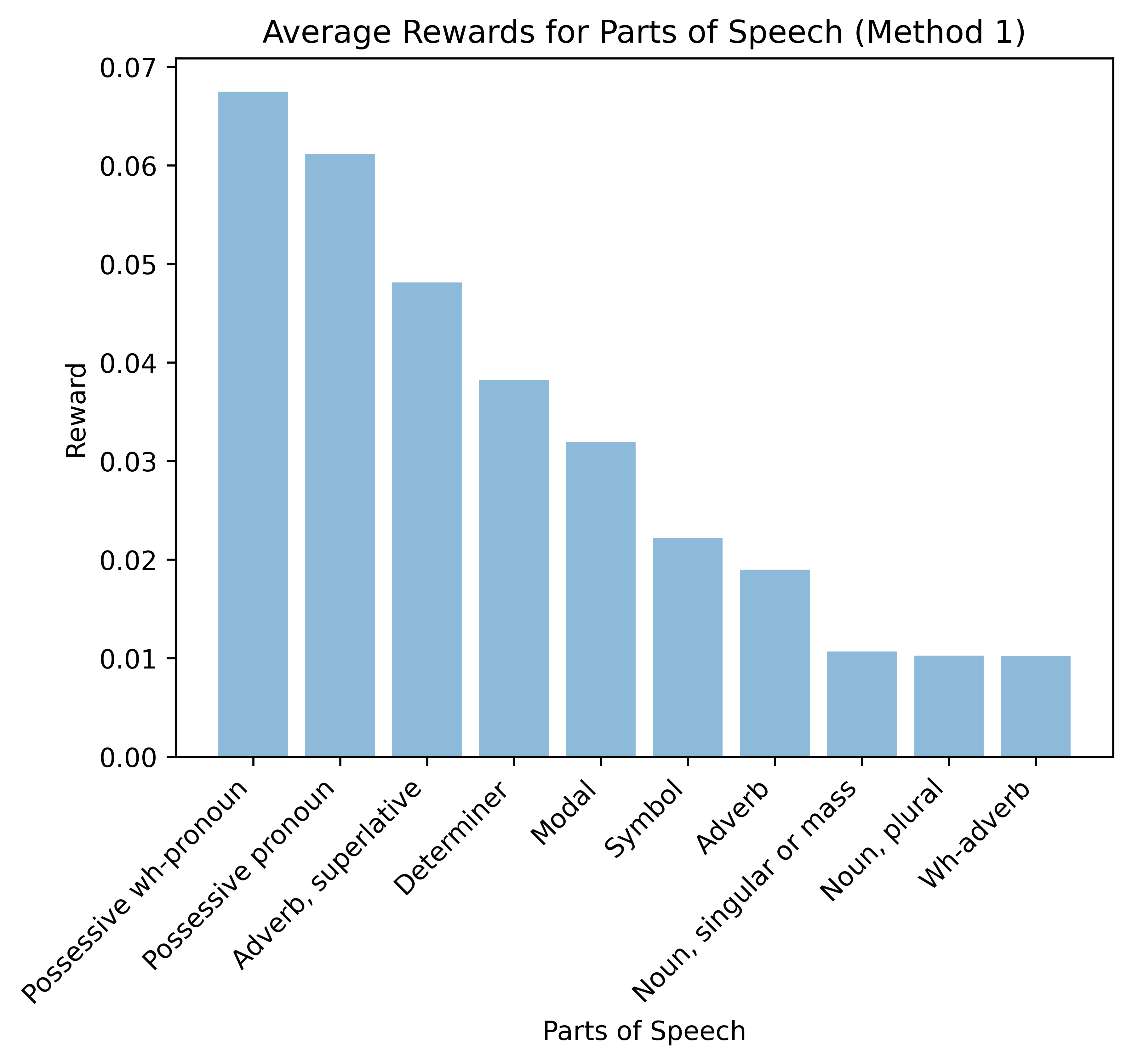}
        \caption{Rewards for parts of speech averaged by Method 1}
        \label{method_1_rewards}
    \end{figure}
    
    \begin{figure}[!t]
        \centering
        \includegraphics[width=2.75in]{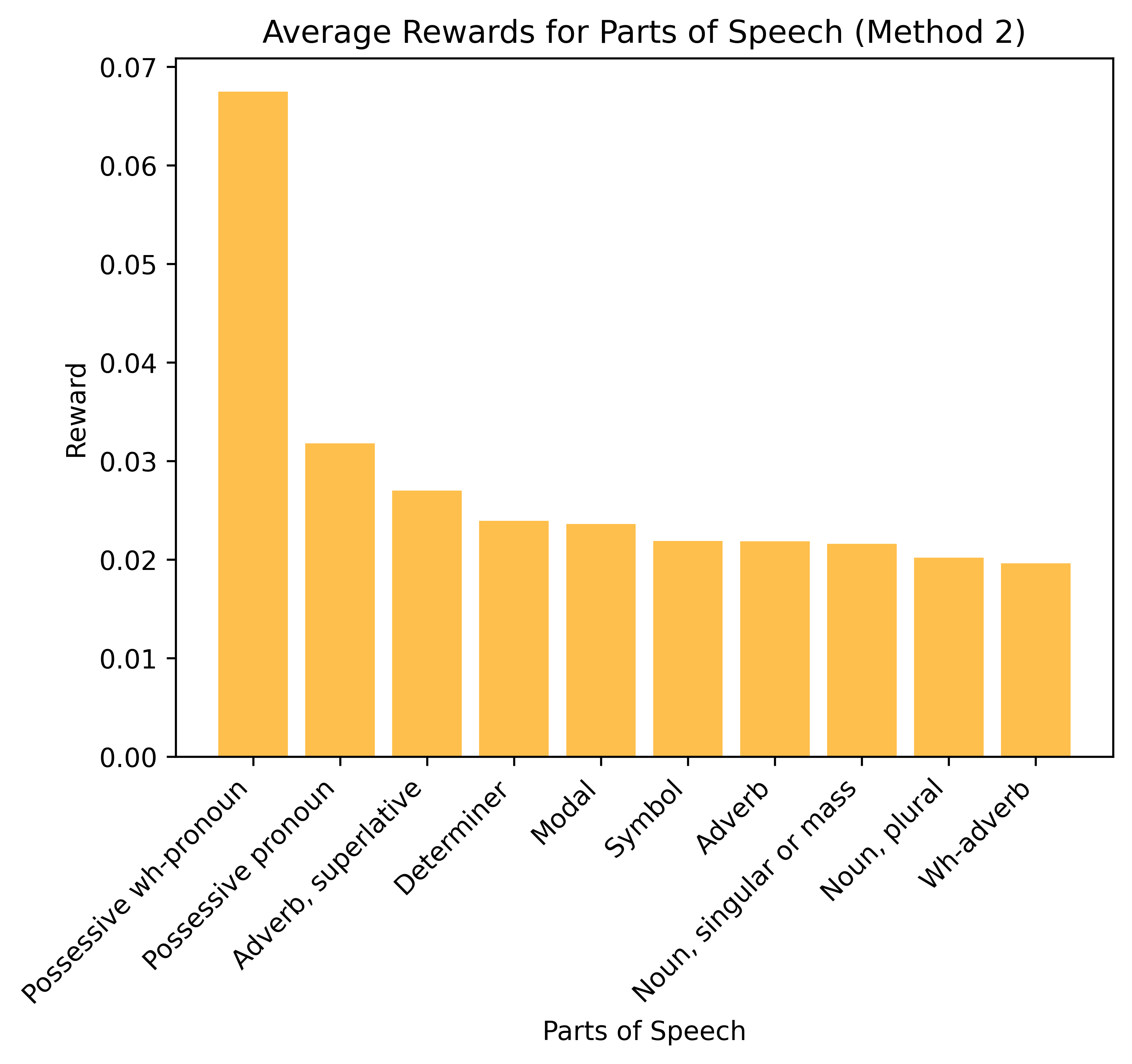}
        \caption{Rewards for parts of speech averaged by Method 2}
        \label{method_2_rewards}
    \end{figure}
    
\section{DISCUSSION}
    \subsection{Model Evaluation} 
        Our expert model achieved near SOTA performance and our
        novice model was able to reproduce the expert’s behavior up to
        the degree that AIRL and IRL usually perform. It is worth
        noting that the goal of this work is to introduce a novel
        interpretability framework for RL-based models rather than
        improving on existing SOTA abstractive summarization
        approaches. To that end, the presented models’ performances
        show the efficacy of our framework. We also would like to
        point out, that, although we chose to train our expert model to
        achieve a high performance and use the expert model for our interpretability experiment, 
        our framework does not require a
        model to achieve any threshold of performance in order to
        explain its decisions. For our task, we found it intuitive and
        appealing to explain the decisions of a well-performing model.
    
    \subsection{Discussion of Results}
        We find \textbf{possesive wh-pronouns, possessive pronouns,
        superlative adverbs, determiners, modal verbs, symbols,
        adverbs, singular nouns, plural nouns and wh-adverbs}, in
        that order, to be the top 10 most rewarding parts of speech
        among words generated by the expert in its summaries. We
        show that the two different methods to calculate average
        rewards only make a difference on the magnitude of average
        rewards but do not make a difference on the ranking of most to
        least rewarding parts of speech.
        We also find a non-zero amount of dependence between the
        number of appearances of a word, the complexity of a word,
        the parts of speech of a word, and the rewards given by the
        discriminator for transitions involving these words. Out of
        these three characteristics examined, the number of
        appearances of a word has the most correlation with the reward
        given by the discriminator for that word, followed by the word
        complexity and its part of speech. 
    
    \subsection{Our Contribution to Interpretability}
        Our framework’s strength lies in its ability to generate
        summaries that explain the model’s behavior over a large set of
        data and be examined in the entire scope of a task. Our method instead
        finds global trends and patterns that underlie the model’s decision-making processes
        over the entire dataset. With our framework, we can make statements that do not
        pertain to single samples of data, but instead broadly describe
        the decisions of a model, such as:
        \begin{itemize}
            \item The model prefers using pronouns over nouns in writing its summaries.
            \item Possessive wh-pronouns are, in this dataset, the most rewarding and most preferred decisions by the model in writing its summaries.
            \item The complexity of a word has slightly more to do with how rewarding a word is than the word’s part of speech. 
                Therefore, word complexity is more influential in the
                model’s decision-making process than parts of speech.
                These global summaries are important because they help
                researchers gain an intuitive, qualitative and/or quantitative
                understanding of the model’s overall decision-making process
                for a specific task.
        \end{itemize}

        For DRL models that are not easily interpretable and tasks
        such as abstractive summarization for which very little amount
        of interpretation is usually provided, our framework can
        provide us with some overarching principles that guide the
        model in its predictions over the samples in the dataset. 
        For models and tasks for which researchers already have an
        intuition or understanding on how a model should work, our framework can be
        applied to gauge the fidelity of the model by extracting patterns and comparing
        them to the established understanding. In either cases, our framework offers practical value by helping researchers and deep learning practitioners gain more confidence and trust in the utilization of their models.
        An IRL-based interpretability framework has yet to be
        adopted by the research community at large. IRL has also not
        been utilized widely for an abstractive summarization task.
        Our framework is the first attempt to utilize IRL, specifically
        Adversarial IRL, to tackle both problems. Through our novel
        framework and our experiments, we showed that IRL can be
        utilized to increase model interpretability and achieve
        promising results for abstractive summarization. In addition to
        abstractive summarization, our framework has the potential to
        be applied for model interpretability in other generative NLP
        tasks.
        We would also like to note that our framework’s output is
        inherently interpretable. Specifically, our framework outputs
        rewards for specific transitions taken by the model, which are
        easily understood. Although not required, with additional
        processing of the output rewards, we can obtain metrics and
        statistics that can be used to create more detailed descriptions
        of the model and thus increase interpretability. 
        
    \subsection{Limitations}
        Popular IRL techniques can only partially achieve the
        performance of the expert model involved in the imitation
        learning process. The performance of our novice is therefore
        bounded by the performance of the expert and the effectiveness
        of IRL as a technique. 
        We believe that although there are
        inherent challenges in training IRL (and GAN) models, the
        advantages of these techniques outweigh the difficulties.
        Our framework currently only helps with the interpretability
        of models in a task if that task can be framed as a
        Reinforcement Learning task.
        This framework would need to be expanded to fit more general tasks before achieving wide
        adoption for broader purposes. In addition, training the novice and the discriminator requires a large amount of data in the form of expert trajectories. Buidling interpretable few-shot learning models and discovering interpretablility methods or frameworks with few-shot learning models remain an ongoing area of research in the field \cite{https://doi.org/10.48550/arxiv.2202.13474, DBLP:journals/corr/abs-1906-05431, Jian2022EmbedHalluc, Jian2022LMSupCon, zhang2021interpretable}. Nonetheless, our framework shows the
        promise of IRL in both the field of XAI as well as the field of NLP. We hope this effort motivates future work and
        experiments with IRL in both fields. 
        
\section{CONCLUSION}
    Increased interpretability of deep learning models leads to
    increased confidence and trust in deep learning models, which
    then lead to the wider adoption of deep learning models in
    practice. Therefore, interpretability frameworks are of
    significant value and interest to the deep learning community.
    In this paper, we introduced a novel interpretability
    framework based on Adversarial Inverse Reinforcement
    Learning (AIRL) that can increase the interpretability of a
    Reinforcement Learning (RL) based model by providing global
    descriptions of a model. We then experimented with the
    framework over an abstractive summarization task. Our models
    achieved promising performance on the abstractive
    summarization task and we showed that our framework can be
    utilized to discover latent patterns and valuable information
    that help us better understand the decision-making process of
    these models.

\section{Acknowledgements}
    We would like to thank Joseph DiPalma, Naofumi Tomita, Wayner Barrios Quiroga, Yiren Jian, Alex DeJournett, Maxwell Aladago and Timothy Daniel Irwin for their feedback on the experimental setup in this work as well as their comments during the writing of this paper.

\newpage
\bibliographystyle{IEEEtran}
\bibliography{IEEEabrv, mybibfile}
\end{document}